\documentclass[journal]{IEEEtran} 
\title{A Survey on Understanding, Visualizations, and Explanation of Deep Neural Networks
}
\author{Atefeh Shahroudnejad$^\dag$\\\vspace{.1in}
$\dag$ Concordia Institute for Information Systems Engineering,  Concordia University, Montreal,~Canada 
}

\usepackage{dblfloatfix}
\usepackage{placeins}
\usepackage{supertabular,booktabs}
\usepackage{flushend}
\usepackage[dvipdfmx]{graphicx}
\usepackage{epsfig}
\usepackage{amsmath}
\usepackage{multirow}
\usepackage{cite}
\usepackage{amsmath}
\usepackage{multicol}
\usepackage{cuted,lipsum}
\usepackage{amsthm}
\usepackage{amssymb }
\usepackage{amsmath}
\usepackage{bm}
\usepackage{cite}
\usepackage{float}
\usepackage{mathrsfs}
\usepackage{amsfonts}
\usepackage{algorithm, algpseudocode}
\usepackage{subfigure}
\usepackage[usenames]{color}
\usepackage[dvipsnames]{xcolor}
\usepackage{tcolorbox}
\usepackage{lipsum}
\usepackage{booktabs,caption}
\usepackage{tabularx}
\usepackage{multirow}
\usepackage{bm}
\usepackage{xcolor,colortbl}
\usepackage{lscape}
\usepackage{nomencl}
\usepackage{amsmath}
\usepackage{mathrsfs}
\usepackage{pgfplots}
\usepackage{tikz}
\usetikzlibrary{arrows,shapes,positioning,shadows,trees}

\usepackage{tabularx}
\usepackage{longtable}
\begin{document}

\date{\today}
\maketitle
\thispagestyle{empty}

\begin{abstract}
Recent advancements in machine learning and signal processing domains have resulted in an extensive surge of interest in Deep Neural Networks (DNNs) due to their unprecedented performance and high accuracy for different and challenging problems of significant engineering importance. However, when such deep learning architectures are utilized for making critical decisions such as the ones that involve human lives (e.g., in control systems and medical applications), it is of paramount importance to understand, trust, and in one word ``explain" the argument behind deep models' decisions. In many applications, artificial neural networks (including DNNs) are considered as black-box systems, which do not provide sufficient clue on their internal processing actions. Although some recent efforts have been initiated to explain the behaviors and decisions of deep networks, explainable artificial intelligence (XAI) domain,  which aims at reasoning about the behavior and decisions of DNNs, is still in its infancy. The aim of this paper is to provide a comprehensive overview on Understanding, Visualization, and Explanation of the internal and overall behavior of DNNs.

\end{abstract}

\section{Introduction} \label{sec:Introduction}

Nowadays, advanced machine learning techniques~\cite{manifold:2016, LSTM_odyssey:2017,Bishop:2006, deep:2016,auto:2018} have strongly influenced all aspects of our lives by taking over human roles in different complicated tasks. As such, critical decisions are being made based on machine learning models' predictions with limited human intervention or supervision. Among such machine learning techniques, deep neural networks (DNNs)~\cite{deep:2016}, such as convolutional neural networks (CNNs)~\cite{ImgN:2014} and recurrent neural networks (RNNs)~\cite{ LSTM_odyssey:2017}\cite{lstm},
have shown unprecedented performance specially in image/video processing and computer vision tasks. However, such deep architectures are extremely complex, full of non-linearities, and also not fully transparent; therefore, it is vital to understand what operations or input information lead to a specific decision before deploying such complex models in mission-critical applications. Consequently, it is undeniably necessary to build the trust of a deep model by validating its predictions, and be certain that it performs as expected and reliably on unseen or unfamiliar real-world data. In particular, in critical domains such as biomedical applications~\cite{medical:2016} and autonomous vehicles (self-driving cars)~\cite{car:2016}, a single incorrect decision cannot be tolerated as it could possibly lead to catastrophic consequences and even threat human lives. To guarantee the reliability of a machine learning model, it is imperative to understand, analyze, visualize, and in one word explain the rational reasons behind its decisions, which is the focus of this article. Despite its importance, explainability of DNNs is still in its infancy and a long road is ahead to open the "black-box" of DNNs. 
Explainability can be fulfilled visually (focus of this work), text based or example based~\cite{Lipton:2016}. Generally speaking, the main goal of an explainable DNN is to find answers to questions of like: What is happening inside a DNN? What does each layer of a deep architecture do? What features is a DNN looking for? Why should we trust decisions  made by a neural network? Having appropriate answers to these questions through development of explainable models results in the following advantages~\cite{Samek:2017}: (i) Verification of the model; (ii) Improving a model by understanding its failure points; (iii) Extracting new insights and hidden laws of the model, and; (iv) Identifying modules responsible for incorrect decisions. To achieve these goals, we focus on the following aspects:
\begin{itemize}
\item \textbf{\textit{Visualizing and Understanding DNNs}}: We start by visualizing features in pre-trained networks to understand the learned kernels (in CNNs) and data structures (in RNNs). We analyze the behavior of DNNs from this point of view and call it Structural Analysis.
\item \textbf{\textit{Explaining DNNs}}: We specifically describe the overall and internal behaviors of DNNs in the explainability concept. Overall behavior (functional analysis) is based on input-output relations, while, internal behavior (decision analysis) tries to dive into mathematical derivatives to describe the decision.
\item \textbf{\textit{Explainability by Design}}: We look into bringing explainability properties into the network design instead of trying to explain a pre-trained network with fixed structure. In other words, we explore DNN architectures that are inherently explainable and discuss how explainability can be achieved by design in general.
\end{itemize}
%
\begin{strip}
\begin{tcolorbox}[colback=green!5,colframe=green!40!black,title=Information Post 1: Explainability of Natural Images and  Illustrative Example]
\small
The information post shows the basics of explainability concept through an illustrative example. The first figure illustrates the black-box concept, where the input is an image and the network prediction is a single word (e.g., face)~\cite{CapsEx:2018}.  As can be seen, such a single output provides no evidence for confirming the truth of predictions or rejecting incorrect predictions without having access to the ground-truth.

\vspace{.25in}
\begin{center}
\includegraphics[width=0.7\textwidth]{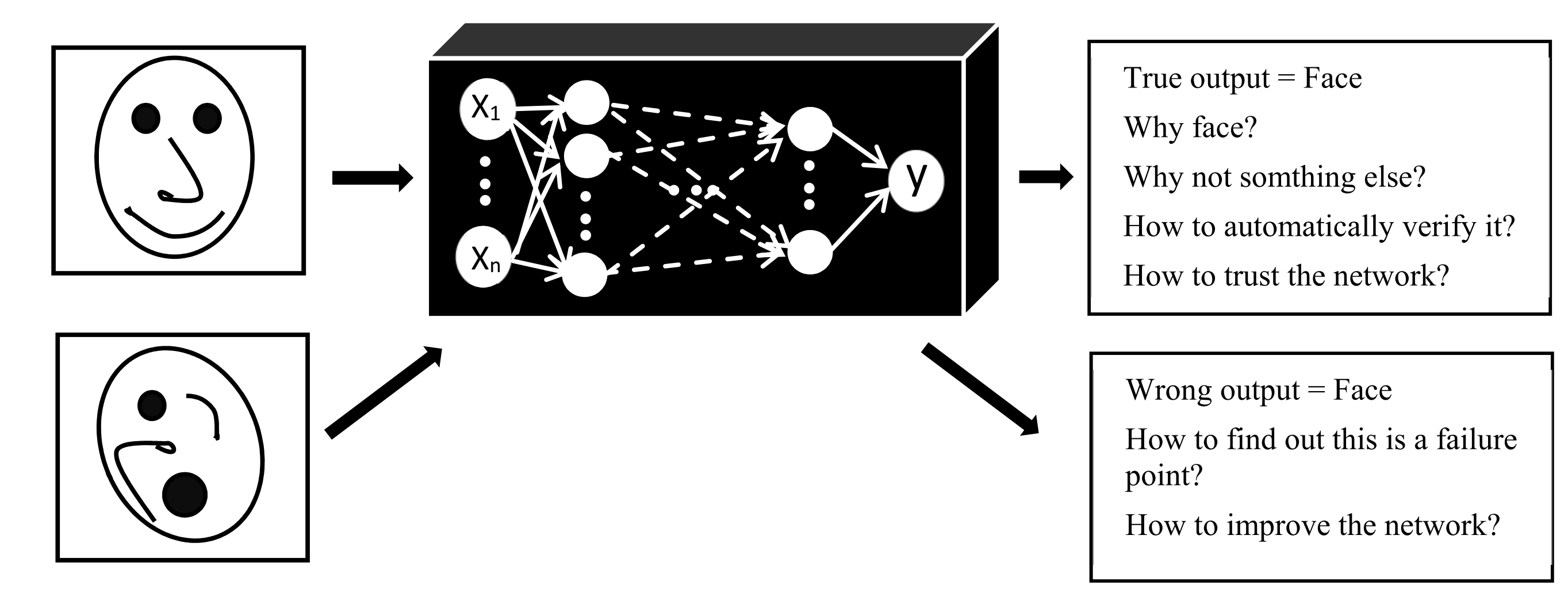}
\end{center}
An example of visual explanation is shown in the second figure. A baseball player has been detected in the image\footnote{https://www.pe.com/2017/04/14/dodgers-clayton-kershaw-rout-zack-greinke-in-7-1-victory/} as the network's decision. The supporting explanation which can be provided here is that there is a player in stadium wearing baseball gloves and holding a baseball ball in his hand.
\vspace{.25in}
\center{\includegraphics[width=0.5\textwidth]{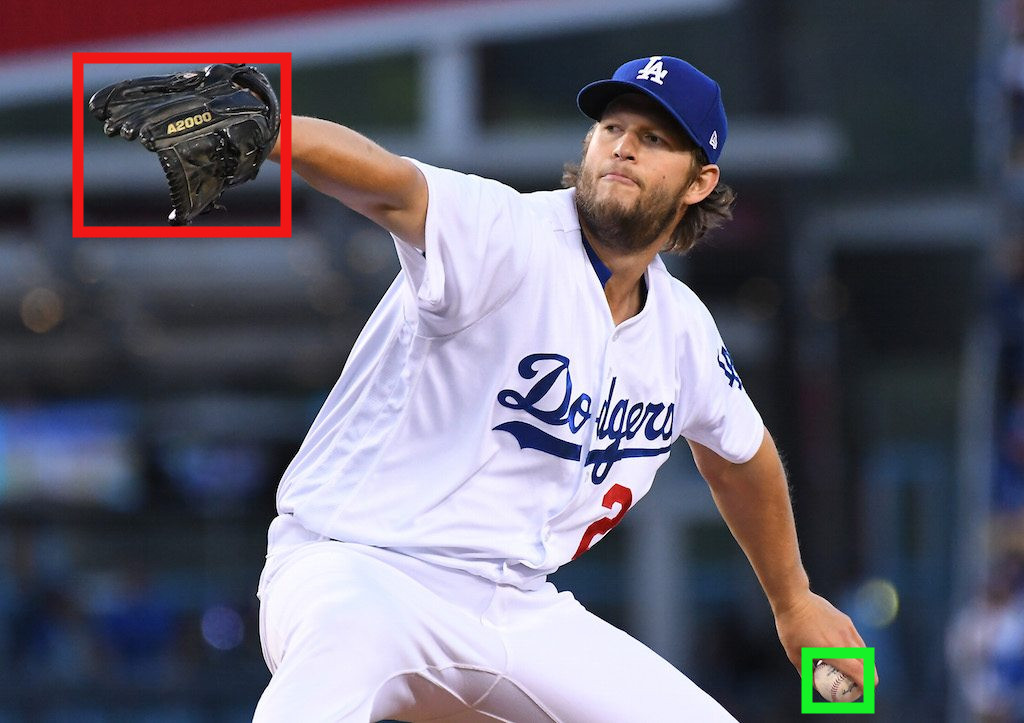}}
\end{tcolorbox}
\end{strip}
\begin{figure*} 
	\begin{tikzpicture}
	[
	basic/.style  = {draw, text width=4cm, drop shadow, font=\sffamily, rectangle},
	root/.style   = {basic, rounded corners=2pt, thin, align=center,
		fill=orange!30},
	edge from parent/.style={->, draw},
	level 1/.style = {sibling distance=6cm, level distance=2cm, basic, rounded corners=6pt, thin,align=center, fill=green!30,
		text width=3cm},
	level 2/.style = {sibling distance= 3cm, level distance= 1.5cm, basic, rounded corners=6pt, thin,align=center, fill=blue!15,
		text width=2cm},
	level 3/.style = {sibling distance=4cm, basic, thin, align=left, fill=pink!60, text width=7.5em}
	]
	
	\node[root] {Underestanding, Visualization and Explanation of DNNs}
	child {node[level 1] (c1) {Structural Analysis} 
		child {node[level 2] (c11) {CNNs}}
		child {node[level 2] (c12) {RNNs}}
	}
	child {node[level 1] (c2) {Behavioural Analysis}
		child {node[level 2] (c21) {Functional Analysis}}
		child {node[level 2] (c22) {Decision Analysis}}
	}
	child {node[level 1] (c3) {Explainability by Design}}
	;

	\begin{scope}[every node/.style={level 3}]
	\node [below of = c11, xshift=20pt] (c111) {First Layer Features};
	\node [below of = c111] (c112) {Intermediate Layers Features};
	\node [below of = c112] (c113) {Last Layer Features};
	
	\node [below of = c3, xshift=15pt] (c31) {CapsNet};
	\node [below of = c31] (c32) {CNN/FF};
	\node [below of = c32] (c33) {semantical object-part filters};
	\end{scope}
	
	\foreach \value in {1,2,3}
	\draw[->] (c11.190) |- (c11\value.west);
	
	\foreach \value in {1,2,3}
	\draw[->] (c3.195) |- (c3\value.west);
	\end{tikzpicture}
	\caption{\footnotesize Overall categorization for understanding, visualization an explanation of deep neural networks}
	\label{fig:category}
\end{figure*}
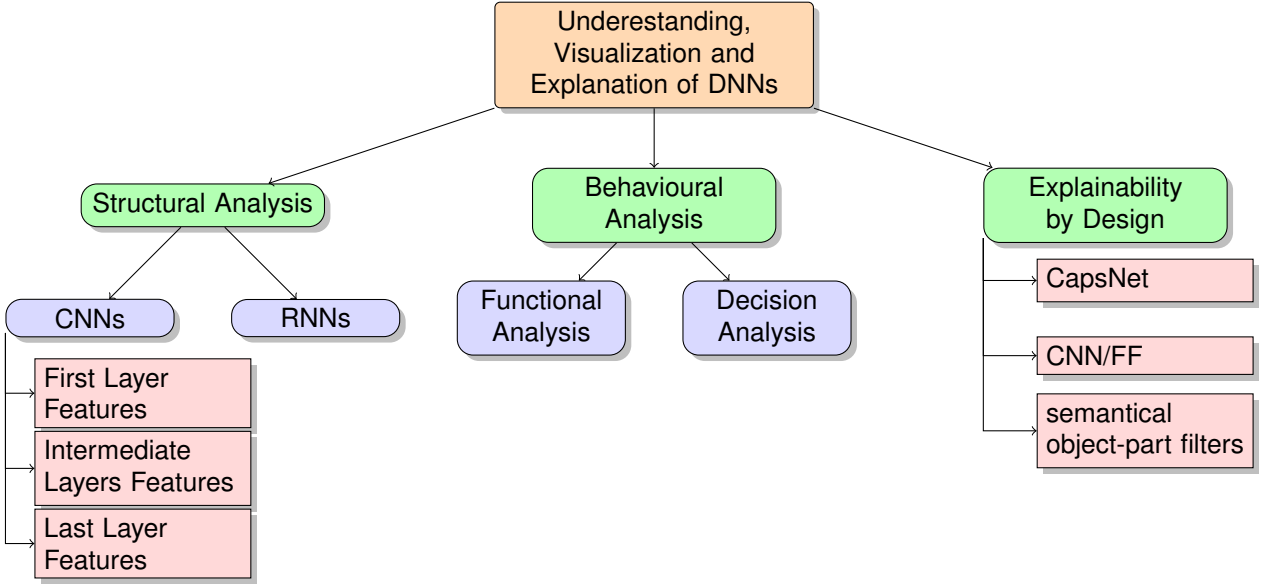

\vspace{.25in}
\noindent
Figure~\ref{fig:category} displays the schematic of these three main categories for understanding, visualization and explanation of deep neural networks.

\textbf{Distinction with Existing Articles}: 
According to the literature, there are few review papers on visualization and explainability of DNNs and none of them covers all the aforementioned aspects together. Moreover, no review on understanding RNNs and explainable designs has been accomplished before. 
For instance, \cite{survey1} only describes activation maximization methods such as~\cite{I8} for understanding DNNs and Layer-wise Relevance Propagation (LRP)~\cite{LRP:2015} for explaining DNNs. The main focus of \cite{survey2} is on four previous works on CNN interpretability which the author has published them recently. paper \cite{survey3} presents a general overview on explaining decision systems; however, no comprehensive review on understanding and visualization  DNNs has been given. 
In contrast, our survey tries to provide a holistic overview on the area.

The rest of this survey is organized as follows: Section~\ref{Str_analysis} reviews methods for visualization and understanding of CNNs and RNNs respectively. Section~\ref{Beh_analysis} investigates explanability based on overall and internal behaviors of DNNs. Section~\ref{Exp_des} illustrates intrinsically explainable models. Finally, section~\ref{sec:Conc} includes conclusion and future works.

\section{Structural Analysis} \label{Str_analysis}
In this section, we try to provide an intuitive answer to the common criticism of deep learning, which is how a deep network makes a decision. Other traditional machine learning techniques such as linear models, decision trees, and random forest are generally easier to interpret because of their simple structures. For example, in linear models, by looking at the weights we can understand how much each input feature effects the decision.
Hence, several methods have been proposed to evidence the meaningful conduct of deep networks.  
In this regard, we investigate CNNs and RNNs as two fundamental categories of DNNs and analyze how information is restructured and represented during the learning process by these networks.

\begin{figure}[b]
	\centering
	\includegraphics[width=0.5\textwidth]{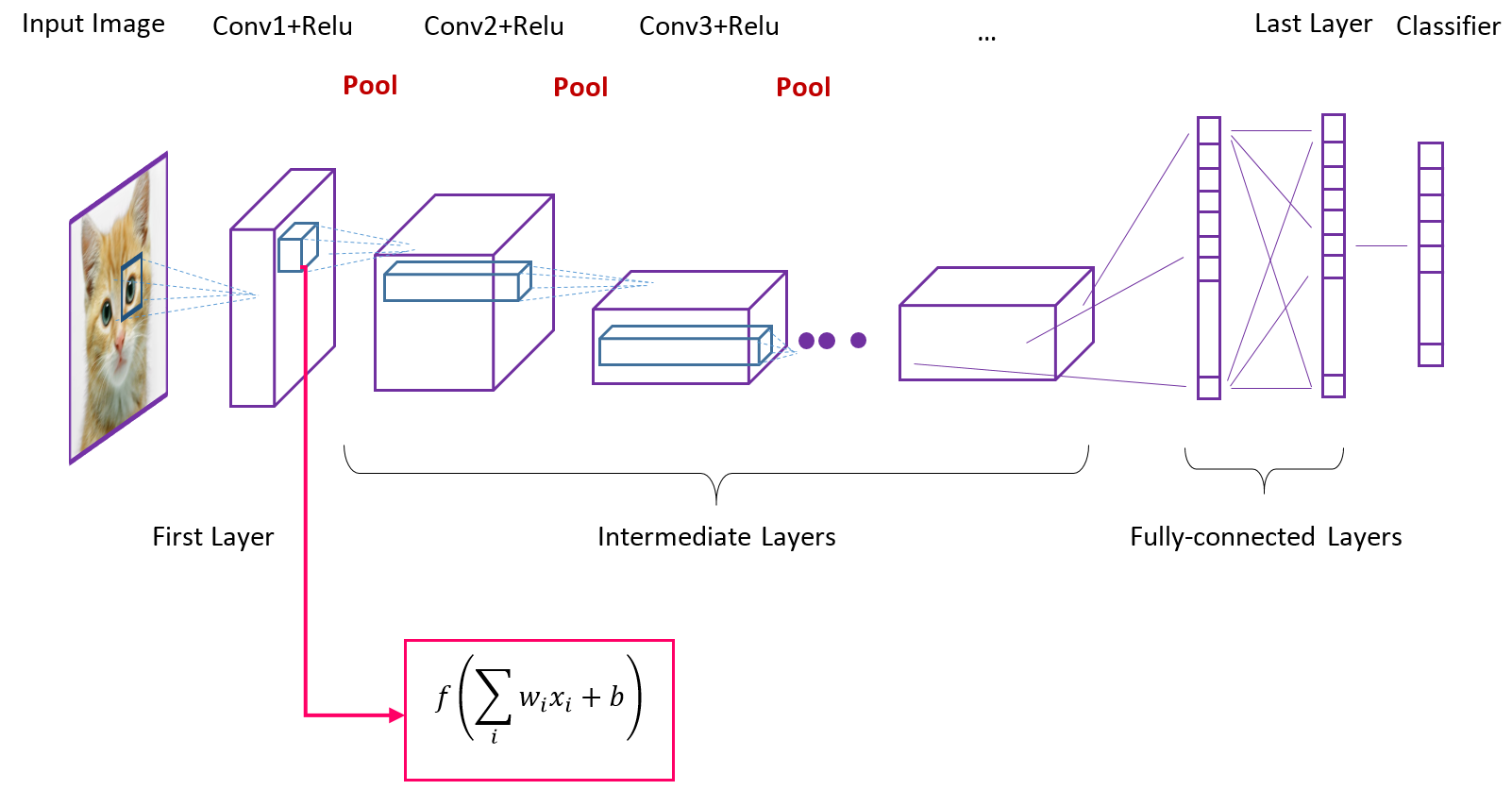}
	\caption{\footnotesize A regular CNN architecture including convolutional layers ("Conv"), pooling layers ("Pool"), and fully-connected layers ("FC").}
	\label{fig:cnn}
\end{figure}
\subsection{Understanding Convolutional Neural Networks}
CNNs contribute significantly to the recent success and interest of DNNs. The most widely used CNN architectures include AlexNet~\cite{ImgN:2014}, VGGNet~\cite{VGG:2014}, ResNet~\cite{Res:2016}, GoogleNet~\cite{Google:2015} and DenseNet~\cite{Dense:2017}, among others. Each convolutional layer in a CNN (Fig. \ref{fig:cnn}) includes several filters such that each of these filters 
will be highly activated by a pattern present in the training set (e.g., one kernel might be looking for curves), which in turn is considered as a "template". As such, interpreting the chain of these learned filters based on different inputs can explain complex decisions maid by CNNs and how the results are achieved. To visualize and understand the learned features of a CNN, we focus on three distinct levels as specified below.

\vspace{.1in}
\noindent
\textbf{\textit{A.1. First Layer Features}}:
Within the first layer, considering that the pixel space is available, filters can be visualized by projection to this space~\cite{ImgN:2014}. The weights of convolutional filters in the first layer of a pre-trained model, indicates what type of low-level features they are looking for, e.g., oriented edges with different positions and angels or various opposing colors. In the next layer, these features are combined together as higher level features, e.g., corner and curves. This process continues and in each subsequent layer more complete features will be extracted.  

\begin{figure}
	\centering
	\includegraphics[width=0.3\textwidth]{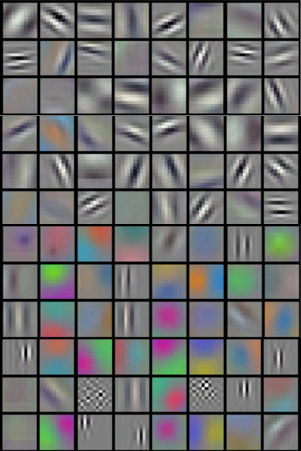}
	\caption{\footnotesize 96 first layer convolutional filters (with size $11\times11\times3$) of pre-trained Alexnet model~\cite{ImgN:2014}.}
	\label{fig:firstL}
\end{figure}

In the first convolutional layer, raw pixels of an input image are fed to the network. Convolutional filters in this layer slide over the input image to obtain inner products between their weights and each image patch which provides the first layer feature maps. 
Since it is a direct inner product between the filters' weights and pixels of input image in the first layer, we can somehow understand what these filters are looking for by visualizing the learned filters' weights. The intuition behind it comes from template matching. The inner product will be maximized when both vectors are the same. For instance, there are $96$ filters of $11\times11\times3$ dimension in the first layer of AlexNet. Each of these $11\times11\times3$ filters can be visualized as a three-channels (RGB) $11\times11$ small image~\cite{ImgN:2014} (Fig.~\ref{fig:firstL}).

As shown in Fig.~\ref{fig:firstL}, there are opposing colors (like blue and orange, pink and green) and oriented edges with different angels and positions. This is similar to human visual system which detects basic features like oriented edges in its early layers~\cite{visual:2015}. Independent of the architecture and the type of training data, the first layer filters always contain opposing colors (if inputs are color images) and oriented edges.
%

\begin{figure*}
	\centering
	\includegraphics[width=1\textwidth]{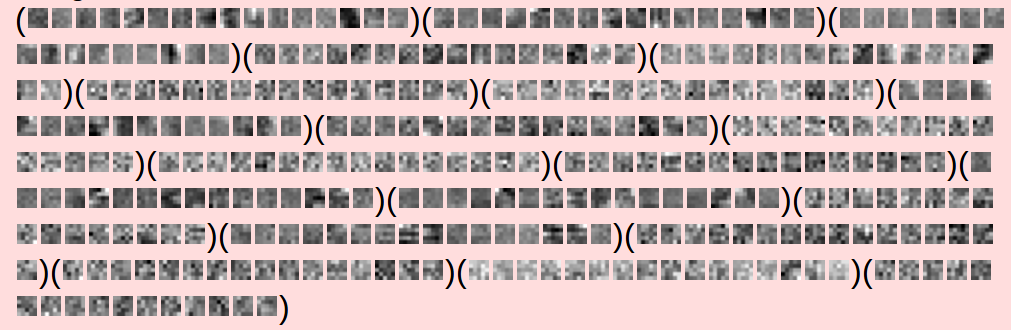}
	\caption{\footnotesize Visualizing filters' weights of second convolutional layer same as the first layer visualization. Here, there are 20 filters with size $5\times5\times16$. This visualization has been taken from ConvNetJS CIFAR-10 demo\protect\footnotemark[\thefootnote].}
	\label{convnetjs}
\end{figure*}

\vspace{.1in}
\noindent
\textbf{\textit{A.2. Intermediate Layers Features}}:

Visualizing features associated with the higher layers of a DNN cannot be accomplished as easy as the first layer discussed above, because projection to pixel space is not straightforward (higher layers are connected to their previous layer output instead of connecting directly to the input image). Therefore, if the same visualization procedure is conducted for the weights of the intermediate convolutional layers (simillar in nature to the approach used for the first layer), we can see which pattern activate these layers maximally; but the visualized patterns are not quite understandable. Since we do not know what previous layer outputs look like in terms of image pixels, we cannot interpret what these filters are looking for. For instance, Fig.~\ref{convnetjs} indicates this weight visualization for the second convolutional layer.

Hence, other alternative approaches are required to discover activities of higher layers. This sub-section investigates methods to understand CNNs based on filters in the intermediate layers and presents their incorporation in a few interesting applications, i.e., feature inversion, style transfer and Google DeepDream\footnote{https://ai.googleblog.com/2015/06/inceptionism-going-deeper-into-neural.html}.

Maximizing filters' activation maps or class scores through back-propagating them to the pixel space is called gradient-based approach. This common visualization approach helps to find out how much a variation in an input image effects on a unit score~\cite{I8}. Also, patterns or patches in input images, which activate a specific unit, will be discovered. 
 
In the following, we demonstrate the gradient-based methods in intermediate layers.

Reference~\cite{I2} visualizes intermediate features by exploring what types of input image patches cause different neurons' maximal activations. Generally, each image patch stimulates a specific neuron. The algorithm is running many images through the pre-trained network and recording feature values for a particular neuron in a particular layer (e.g. Conv5). Then image patches in dataset corresponding to the parts of that neuron which has been maximally activated, are visualized. 
Here, to detect most responsible pixels for higher activation of a neuron through deconvolution process, the paper proposes Guided Back-propagation instead of regular back-propagation to have much cleaner result images. In Guided Back-propagation since only positive gradients are back-propagated through ReLUs, only positive influences are tracked throughout the network.  
The intuition behind this projection down to pixel space is obtaining saliency map for that neuron (same as saliency map for a class score described in section~\ref{sec:FA}) which is accomplished by computing the gradient of each neuron value with respect to the input image pixels.

\footnotetext[\thefootnote]{https://cs.stanford.edu/people/karpathy/convnetjs/demo/cifar10.html}

However, for more understanding it is necessary to know what type of input \textbf{in general} maximizes the score of an intermediate neuron (or a class score in case of section~\ref{sec:FA}) and causes this neuron to be maximally activated. For this purpose, in a pre-trained network with fix weights, gradient ascent is applied iteratively on pixels of a zeros-(or uniformly) initialized image which is called synthesized image and it is updated accordingly. Here, a regularization term is needed to prevent pixels of the generated image from overfitting to some specific features of the network. The generated image should maximally activate a neuron value (or a class score) while appearing to be a natural image. Hence, it should have the statistics of natural images and a regularization term ($R_\theta(X)$) shall force it to look almost natural. A common regularizer is the $L2$ norm of synthesized image ($\|X\|^2$), which penalizes extremely large pixel values. But through this regularization, the quality of generated results is not enough and therefore it needs to be improved by improving regularization term instead of using just $L2$ one.

In this regard, paper~\cite{I1} uses two other regularizers in addition to $L2$ norm as natural image priors which are: the Gaussian blur to penalize high frequency information; setting pixels with small values or low gradients to zero during the optimization procedure (Eq.~\eqref{reg}). Using these regularizers makes generated results look more impressive and much clear to see.
\vspace{.05 in}
\begin{equation}\label{reg}
X^*=\operatorname*{argmax}_X(f_i(X)-R_\theta(X))
\end{equation}  
~\cite{I1} generally proposes an interactive deep visualization and interpreting toolbox whose input can be an image or a video. The toolbox displays neurons' activations on a selected layer of a pre-trained network and it gives the intuition of what that layer's neurons are doing. For example, units in the first layer response strongly to sharp edges. On the other hand, in higher layers (e.g., Conv5) they response to more abstract concepts (e.g. cat's face in cat image). The toolbox also shows synthesized images to produce high activation via regularized gradient ascent as mentioned before. It means that the neuron fires in response to a specific concept. By feeding all training set images and selecting top ones that activate the particular neuron most and also computing pixels which are most responsible for the neuron's high activation through backward deconvolution process, we can concretely analyze that neuron.

In references~\cite{I6-1,I6}, the weights of the output layer (e.g., class scores) are back-propagated to the last convolutional layer to determine the importance of each activation map in the decision. Moreover, GradCAM~\cite{I6-1} and Guided Back-propagation~\cite{I2} can be combined together through element-wise product to result more specific feature importance~\cite{I6-1}.

Another approach to improve visualization is considering extra features. In this regard, paper~\cite{I10} tries to explicitly consider multiple facets (feature types) of a neuron in the optimization procedure rather than assuming only one feature type. "Multiple facets" refers to different types of images in the dataset where all of them are in the same class and convey same concept (e.g. different colors of an object). To visualize a neuron's multiple facets, the algorithm first uses k-means to cluster all different images which highly activate the neuron. Each cluster is considered as a different facet and the mean of its images is used to initialize a synthesized image for the facet. Then, activation maximization procedure of the neuron is applied to all facets' synthesized images. 
Paper~\cite{I11} proposes a deep image generator network (DGN) as a learned natural image prior for synthesized image in activation maximization framework. The generator is the generator part of GAN, which is trained to invert extracted feature representation by the encoder such that the synthesized image has features close to real images and the discriminator can not distinguish it from real ones. The final objective is optimizing latent space representation (instead of pixel space) to generate a synthesized image which maximizes the activation of determined neuron. By adding such additional priors toward modeling natural images, we can generate very realistic images at the end which indicate what the neurons are looking for.  

In the following, we briefly refer to the most popular applications of understanding intermediate filters.

\textbf{Applications}:
The idea of synthesizing images using gradient ascent is quite powerful and it can be used for generating adversarial examples to fool the network. By choosing an arbitrary image and an arbitrary class, the algorithm starts modifying the image to maximize the selected class score and repeats this process until the network is fooled. The results are much surprising because they look same as original images but they are classified as selected classes. 
Google has also proposed another application named DeepDream, which creates dreamy hallucinogenic images. It again indicates what features the network are looking for. DeepDream tries to amplify some existing features which have been detected by the network in an input image through maximizing neuron activations in a layer. After running an input image through the pre-trained network up to the chosen layer, the gradient of that layer (which is equal to its activation (Eq.~\eqref{deepdream})) is computed with respect to input image pixels and the input image is updated iteratively to maximize this layer activation. when we select lower layers, many edges and swirls pop up in the image. On the other hand by selecting higher layers, some meaningful objects appear in the image. 
\vspace{.05 in}             
\begin{equation}\label{deepdream} 
I^*=\operatorname*{argmax}_I\sum_{i}f_i(I)^2 
\end{equation} 


\textbf{Feature inversion} concept indicates what types of image elements are captured at different layers of the network.
In feature inversion an image runs through the pre-trained network and its feature values are recorded. After reconstructing that image from the feature representation of different layers, regarding what reconstructed images look like, they give some sense about how much information have been captured in these feature vectors~\cite{I4}. For instance, if the image is reconstructed based on first layers' features (e.g. ReLU2  in VGG-16), we can see the image is well reconstructed which means that less information is thrown away at these layers. in contrast, in deeper layers although the general spatial structure of the image is preserved, a lot of low level details such as exact pixel values, colors and texture are lost which means that in higher layers the network is throwing away these kind of low level information and instead it focuses on keeping around more semantic information.     
It synthesizes a new image ($X$) by matching its features ($\Phi(X)$) back to that recorded features ($\Phi_0$) which are computed before for that layer. It minimizes the distance between them using 
gradient-ascent and total-variation regulariser (Eq.~\eqref{fI}).
\vspace{.1 in} 
\begin{equation}\label{fI}
X^*=\operatorname*{argmin}_{X\in \Re^{H\times W \times C}} \|\Phi(X)-\Phi_0\|^2+\lambda R_{V^\beta}(X),
\end{equation}
\vspace{.05 in}
where the result image ($X^*$) has the most similar features to the recorded ones. $\lambda$ is the regularizer coefficient, and total variation regularizer ($R_{V^\beta}$) is another natural image prior which penalizes differences between adjacent pixels vertically and horizontally, which makes the generated image smoother. However, this natural image objective is hardly achieved in practice. Therefore, the authors improved their method using natural pre-images and more appropriate regularizers~\cite{I5}.

Feature reconstruction loss is also used in single image super-resolution problem~\cite{style2:2016}. It transfers semantic knowledge in this way and therefore it can more precisely reconstructs fine details and tiny edges rather than previous methods such as Bicubic and SRCNN.

\textbf{Texture synthesis} problem is generating a larger piece of a texture from a small patch of it. Nearest neighbor is one classical approach for texture analysis which works well for simple textures. It generates one pixel at a time based on its neighborhoods which have been generated before and copying the nearest one. But for more complex textures, neural network features and gradient-ascent procedure are required to match pixels. Neural network synthesis uses the concept of gram matrix~\cite{texture:2015}.
It runs the input texture through a pre-trained CNN and records convolutional features at each network layer ($C\times{H}\times{W}$ feature tensor at each layer). 
According to Eq.~\eqref{gram}, the outer product between every two of these $C$-dimensional feature vectors ($F$) in each layer indicates second order co-occurrence statistics of different features at that two points. 
\vspace{.05 in}
\begin{equation} \label{gram}
G_{ij}^l=\sum_{k}F_{ik}^l F_{jk}^l
\end{equation}
Gram matrix ($G$) gives the sense that which features in feature map tend to activate together at different spatial positions. The average of all $C\times{C}$ feature vectors pairs in each layer creates a gram matrix for that layer as a texture descriptor of input image. There is no spatial information in gram matrix because this average operator.    
Using these texture descriptors, it is possible to synthesize a new image with similar texture of original image through a gradient-ascent procedure. It reconstructs gram matrix (texture descriptor) instead of whole feature map reconstruction of the input image.
If we use of an artwork instead of input texture in texture synthesis algorithm (by maximizing gram matrices), we will see the generated image tries to reconstruct pieces of that artwork.

\textbf{Style transfer} is the combination of feature inversion and texture synthesis algorithms. It defines a global reconstruction loss on them ($\ell_{total}$). In style transfer~\cite{style:2016}, there are two input images called content and style image. The output should looks like to content image with the general texture of style image. Therefore, the algorithm generates a new image through minimizing the feature reconstruction loss of the content image ($\ell_{content}$) and the gram matrix loss of the style image ($\ell_{style}$). 

\begin{equation}\label{style}
\ell_{total}=\alpha\ell_{content}+\beta\ell_{style}\\
\end{equation}
\vspace{.05 in}        
Here, using feature inversion and texture synthesis together in Eq.~\eqref{style} brings more control on what the generated image will look like compared to DeepDream. 
By considering a trade-off between content and style loss ($\alpha$ and $\beta$ multipliers), we determine how much the output matches to content and style. By resizing style image before applying the algorithm, it indicates the features scale for reconstruction from style image. Style transfer can also be done with multiple style images by taking a weighted average of different styles' gram matrices at the same time. 
Style transfer can be combined with DeepDream by considering three losses: style loss, content loss, and DeepDream loss which maximizes neuron activations in a layer.    
Since style transfer consists of many iteratively forward and backward passes through the pre-trained network, it is very slow and needs a lot of memory. Hence, paper~\cite{style2:2016} proposes a real-time style transfer approach which is fast enough to be applied on video. The method trains a feed-forward image transformation network on a dedicated style image. This network is fed by a content image and its output is directly stylized image. Therefore, there is no need to compute several optimization procedure for each synthesis image like~\cite{style:2016}. 
In the training phase, the weights of feed-forward network are updated through minimizing the global loss, which is combination of content and style losses as before. This process iterates for tens of thousand content images. In test time, there is no need to feature extraction and loss network. It is enough to feed content image to the trained feed-forward network.  
Paper~\cite{style3:2016} also improves the speed of style transfer in a similar way. It consists of two networks : a feed-forward generative network which is trained and finally used for stylizing images; and a pre-trained descriptor network which calculates losses for each training iteration. They also improves their results by replacing batch normalization with instance normalization in generator network~\cite{style5:2016}.   
These fast style transfer networks are computationally expensive because a separate feed-forward network should be trained for each style. To tackle this limitation, paper~\cite{style6:2017} trains just a single feed-forward network same as ~\cite{style2:2016} for capturing multiple styles in real-time using conditional instance normalization. This approach allows all convolutional weights of the network are shared across all those styles. The only difference for modeling each style is that separate shift and scale parameters are learned after normalization for that style. The network can also blends styles in test time. 

Paper~\cite{I9} defines generality and specificity concepts for layer features and quantifies how much neurons of each layer are general or specific. The first layer features are completely general because they are in a lower level. On the contrary, last layer features are totally specific to a particular task. Feature transition from general to specific can be used for determining transfered layers when fine-tuning a network for a specific target.

   
\vspace{.1in}
\noindent
\textbf{\textit{A.3. Last Layer Features}}:
The last layer produces a summary feature vector of the input image. This sub-section will investigate these high-dimensional feature visualization in CNNs. Generally speaking, there are fully-connected layers and a classifier at the end of CNNs. The classifier provides predicted scores of each class category in the training dataset. The fully-connected layer before the classifier provides feature representation vector of input image, which is fed to the classifier for making final decision. For example in AlexNet these feature vectors have 4096 dimension.  

\begin{figure}[t!]
	\centering
	\includegraphics[width=0.49\textwidth]{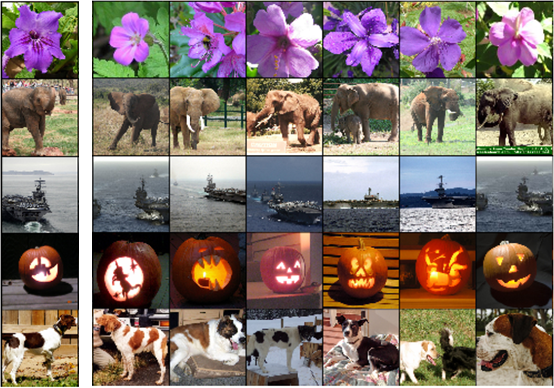}
	\caption{\footnotesize Top six nearest training images which their produced feature vectors in the last layer have smallest Euclidean distance form the feature vector of corresponding test image in the left~\cite{ImgN:2014}.}
	\label{fig:lastL}
\end{figure}

In this regard, paper~\cite{ImgN:2014} feeds images in the datasets to the pre-trained CNN and after running them through the network, it collects their feature vectors of last fully-connected layer. Then, it uses nearest neighbors method to interpret the last hidden layer. If feature vectors of two images have small Euclidean distance, it means that the network considers them similar. Nearest neighbors in feature space (rather than pixel space) considers the similarity of images in terms of semantic content. As we can see in Figure~\ref{fig:lastL}, the pixels in the test images are kind of different from those in their nearest neighbors in training images, but they are in the same feature space which represents the semantic content of those images.

\begin{figure}[t!]
	\centering
	\includegraphics[width=0.49\textwidth]{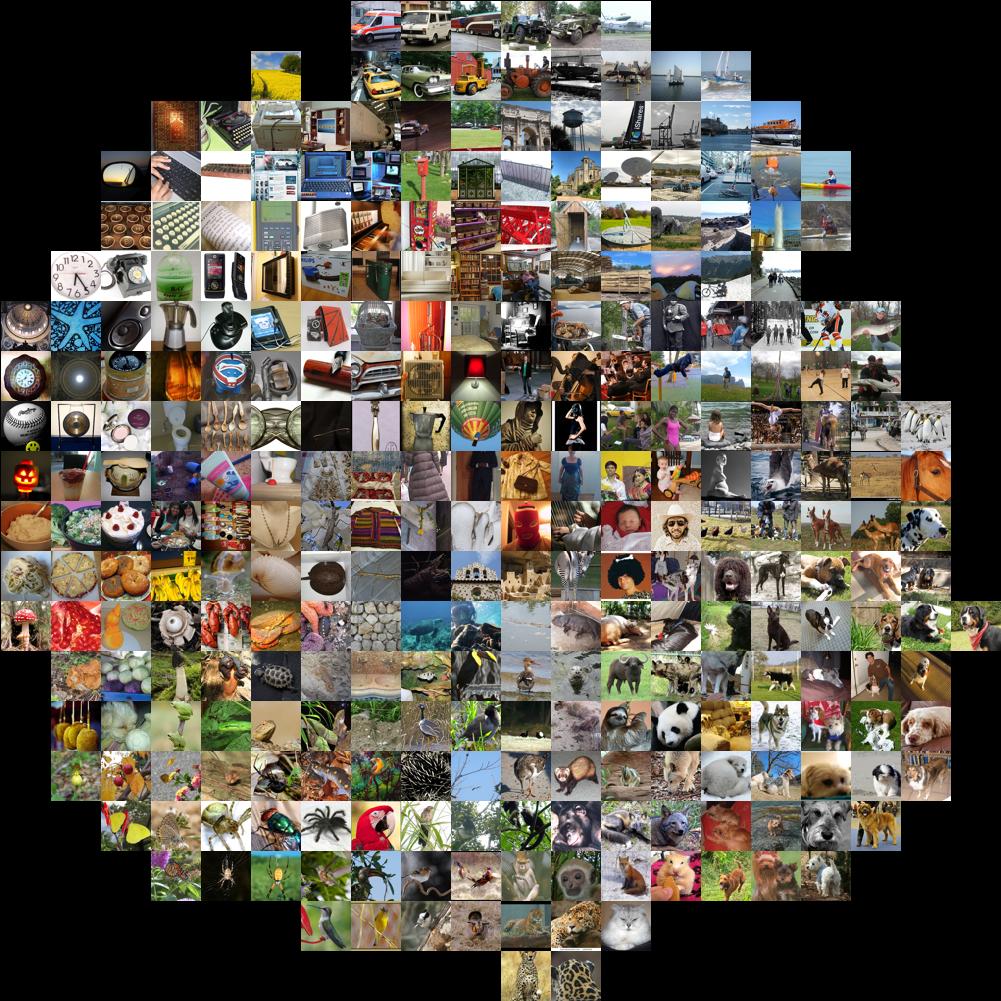}
	\caption{\footnotesize t-SNE visualization for the last layer features of the pre-trained ImageNet classifier   \protect\footnotemark[\thefootnote]. Each image has been displayed at its corresponding position in two-dimensional feature space.}
	\label{fig:tsne}
\end{figure}

Another approach for visualizing the last layer is by dimensionality reduction. The most important aim in this category is to preserve the structure of high-dimensional data in low-dimensional space as much as possible. It makes the feature space visualizable by reducing feature vectors' dimensions to two or three dimensions which can be displayed. For instance, Principle Component Analysis (PCA) is one of the traditional linear dimensionality reduction techniques. However, when high-dimensional data stand close to low-dimensional, it is usually impossible to keep similar data-points close to each other using linear projections. Hence, non-linear methods are more suitable for high-dimensional data because they can preserve the local structure of data very well. In this regard, paper~\cite{L2} introduces t-Distributed Stochastic Neighbor Embedding (t-SNE) dimensionality reduction technique for high-dimensional data visualization. It models similar data-points by small pairwise distance and dissimilar ones by large pairwise distance. Therefore, t-SNE captures data local structure while tracing global structure (e.g. clusters occurrence at different scales). By applying t-SNE technique to the obtained feature vectors of dataset images from the last layer of a trained network, high-dimensional feature space is compressed to two-dimensional one. Therefore, we can see what types of images are in the same location in the t-SNE representation (two-dimensional coordinate) and also have some sense about the geometry of the learned features as well as certain semantic concepts (e.g. different types of flowers are in same cluster). Figure~\ref{fig:tsne} illustrates t-SNE visualization for the last layer features of the pre-trained ImageNet classifier.             

\footnotetext[\thefootnote]{https://cs.stanford.edu/people/karpathy/cnnembed}
\subsection{Understanding Recurrent Neural Networks}
\begin{figure}[t!]
\centering
\includegraphics[width=0.5\textwidth]{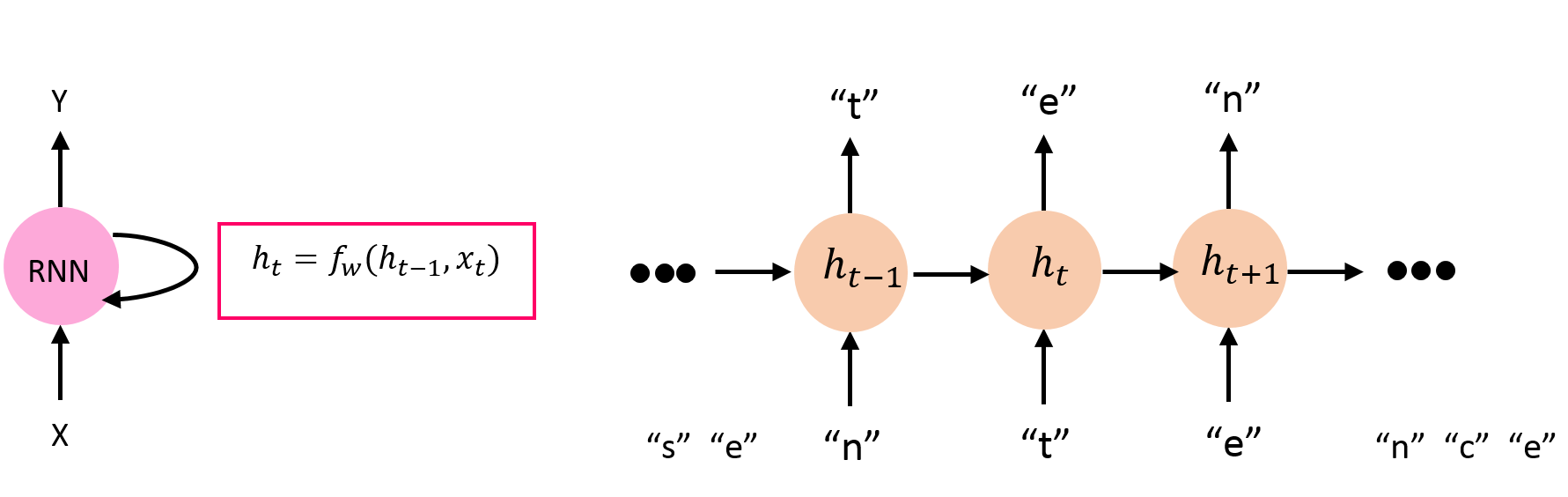}
\caption{\footnotesize A simple RNN architecture unrolling.}
\label{fig:rnn}
\end{figure}

RNNs are considered as exceptionally successful category of neural network models specially in applications with sequential data such as text, video and speech. Commonly used RNN architectures include VanillaRNN (Fig. \ref{fig:rnn}), Long Short-Term Memory (LSTM), and Gated Recurrent Unit (GRU), which have been applied in various applications such as machine translation~\cite{rnn_app1}, language modeling~\cite{rnn_app2}, text classification~\cite{rnn_app3}, image captioning~\cite{rnn_app4}, video analysis~\cite{rnn_app5} and speech recognition~\cite{rnn_app6}. Although RNNs' performances are impressive in these areas, it is still difficult to interpret their internal behaviors (what they learn to capture) and also understand their shortcomings.


Analyzing RNNs' behavior is most challenging because: First, hidden states perform like memory-cells and store input sequence information. Thousands of these hidden states should be updated with millions parameters of nonlinear functions, when an input sequence (e.g. text) is fed to the network. Therefore, due to number of hidden states and parameters, interpreting RNNs' behavior is difficult. Second, There are some complex sequential rules in sequential data which are difficult to be analyzed (e.g. grammar and language models embedded in texts). Third, there are many-to-many relationship between hidden states and input data which effects on understanding hidden states information~\cite{RNN2}. Fourth, in contrary to CNNs which their input space is continuous, RNNs are usually applied on discretized inputs such as text words. Moreover, each hidden unit gradient depends on its previous hidden unit. Hence, gradient-based interpretation methods such as activation maximization are barely applicable on intermediate hidden units.    

In this regard, some research studies have tried to address these challenges and interpret the hidden state units' behavior in different RNN models by monitoring representation of these units changes over time. paper~\cite{RNN1} introduces a static visualization method. It uses character-level language modeling (i.e., the input is a characters sequence and pre-trained RNN is supposed to predict next character) as an interpretable testing platform to understand learned dependencies and activations in RNNs. They demonstrate that there are some interpretable hidden state units in language model, which can be understood as language properties. They select a single hidden unit and visualize its value as a heatmap on data. For instance, one unit is sensitive to word positions in a line; the other one illustrates the depth of an expression; some other units turn on inside comments and quotes, indentation and conditional statements (e.g. if statement in a block of code). 

However, studying the changes of hidden units to discover interpretable patterns includes a lot of noise. In this regard, paper~\cite{RNN_lstmvis}
generalizes the idea of~\cite{RNN1} and proposes a visualization interactive tool to identify hidden units dynamics in LSTM and conduct more experiments. This visualization tool allows users to selectively trace and monitor changes of local unit vectors over time by specifying an input range and determining a hypothesis about hidden unit properties to capture a pattern. These local unit changes are adjustable to find similar phrases with this pattern in large datasets. For a hypothesis (e.g. selecting noun phrases in language modeling), the number of activated hidden units at each time step is encoded in a match-count-heatmap and all founded matched phrases are also provided. 
Moreover, there are extra annotations as ground-truth which helps users to confirm or reject the hypothesis. However, many hidden units still exist, and the semantics behind them are not clear, which makes it difficult to understand their dynamics (which patterns they capture). 
Moreover, formulating a hypothesis requires some insights about the values of hidden units over time.      

Paper~\cite{RNN_NLP} proposes a gradient-based saliency technique for visualizing local semantics and determining the importance of each word in an input text and its contribution in the final prediction (e.g. text classification). Values of hidden units' vectors are plotted over a sequence. But this plot is might be unscalable when there are a large number hidden units in the model. The saliency heatmaps for same sentences which are fed to three different trained RNN models (VanillaRNN, LSTM and Bi-Directional LSTM) confirm that LSTM architectures present clearer emphasis on substantial words and decrease the effects of irrelevant words. In the other word, LSTMs can learn to capture important words in sentences. However, it is critical to go beyond the words and focus on structures and semantics learned by these networks. 
Considering the scalability deficiency in~\cite{RNN_NLP}, paper~\cite{RNN2} uses bi-graph partitioning and co-clustering visualization to model the relation between hidden units and inputs (e.g. text words) for more data structure investigation in hidden units.  
For individual hidden unit evaluation, the function of each single hidden unit $\textbf{\textit{i}}$ ($h_i$) is understood using exploring its most salient words which are defined based on model's response to a word $\textbf{\textit{w}}$ at time $\textbf{\textit{t}}$. Eq.~\eqref{ss} indicates the relation between $h_i$ and word $\textbf{\textit{w}}$. Most salient words for unit $\textbf{\textit{i}}$ have larger $|s(w)_i|$. 
\begin{equation}\label{ss}
s(w)_i=E(\Delta h_i^t|w_t=w),
\end{equation}
\begin{align*}
\Delta h_i^t=h_i^t-h_i^{t-1},
\end{align*}
where $\Delta h_i^t$ is considered as response of $h_i$ to word $\textbf{\textit{w}}$ at time $\textbf{\textit{t}}$ .
For sequence evaluation, glyph-based visualization is used to interpret how hidden units change dynamically and behave over a sequence input.  

Paper~\cite{RNN_attention} have added an attention mechanism to the decoder part in machine translation context (sequence-to-sequence model which can be considered as a combination of many-to-one (encoder) and one-to-many (decoder)). This mechanism assigns attention weights to each hidden state unit in the decoder part and it allows decoder to select parts of the input text to pay attention on them. Here, attention weight is defined as a probability of translation a target word from an input word which represents the importance of corresponding hidden unit. Therefore, the model automatically looking for the phrases in the input text which are relevant to next word prediction. Paper~\cite{RNN_attention2} defines an omission score to quantify the contribution of each word in final representation. The score is computed based on the difference between sentence representations with and without that word. They also try to learn lexical patterns and grammatical functions which take more semantic information through changing each word position and analyzing the resulting representations. For determining related hidden units to a specific context, mutual information has been used. 
Paper~\cite{RNN_erasure} tries to analyze decisions made by a RNN through monitoring the effects of removing the following items on the network's decision: input words, word array dimensions and hidden state units as different parts of representation. They apply reinforcement learning to find least number of words which should be removed for changing model's decision.             

\section{Behavioral Analysis} \label{Beh_analysis}
The main goal of a DNN's explanation is analyzing its behavior and answering this question: Why should we trust this neural network's decision?
For this desired, explainability requires extra qualitative information about how the model reach a specific decision. 
In this section, we study explainability based on overall and internal behaviors of DNNs. Functional Analysis tries to capture overall behavior by investigating the relation between inputs and outputs. On the other hand, Decision Analysis sheds light on internal behavior through probing internal components rolls.
\subsection{Functional Analysis}\label{sec:FA}
Generally speaking, approaches belonging to the functional analysis category provide explainability by considering the whole neural network as a non-opening black-box and trying to find the most relevant pixels 
for a specific decision regarding the input image. In other words, network's operation is interpreted through discovering the relation between inputs and their corresponding outputs.
This sub-section investigates such explainable methods.

One approach is applying a sensitivity analysis for interpreting classifiers' predictions which it means changing the input image and observing the consequent changes in the results. For instance, paper~\cite{I2} approximates how much each part of the input image is involved in the decision of network's classifier. The algorithm blocks out different regions of an input image with a sliding gray square and then it runs these occluded images through the network and displays their probabilities for correct class using a heatmap. When the blocked region corresponds to determinative part of the image (it is important in true decision), the correct class probability will decrease.
Paper \cite{pred_diff:2017} also applies a similar approach to obtain the heatmap. Here, instead of blocking out, each intended region is sampled from the surrounding larger window (except the region area). The difference between two class probabilities (original and sampled patch) indicates the importance of that region. 
A gradient-based method has been proposed in ~\cite{SA:2010,I3} which determines pivotal pixels in a non-linear classifier's decision using saliency map. They compute gradient of the predicted class score $S_c$ with respect to the input image pixels ($\partial S_c/\partial I$); which means that how much classification score will change by stumbling each pixel. However, since the resulting saliency maps are noisy and visually diffused, SmoothGrad~\cite{smoothgrad} has been proposed to suppress the noise by averaging over saliency maps of noisy copies of an input image. Moreover, to address probable gradient saturation problem and also capture important pixels more precisely, interior gradients~\cite{IG} integrates gradients of the scaled copies of original input instead of considering only the input's gradient. 

Another approach, named Local Interpretable Model-Agnostic Explanations (LIME), assumes that all complex models are locally linear. Hence, it explains predictions by approximating them with an interpretable simple model around several local neighborhoods~\cite{Marco:2016}. LIME creates fake dataset by permuting original observations and then measures similarity between those generated fake data and the original ones. These new data are fed to our neural network and their class predictions are calculated. Then, top best-describing features which give maximum likelihood of the predicted class, are selected and a simple model (e.g. linear model) is fitted to data based on it. The weights of this model are considered as an explanation for local behavior of the complex model.
Despite LIME with model locally linear assumption, Anchor approach~\cite{anchor:2018} can capture locally non-linear behaviors by applying interpretable if-then rules (i.e., Anchors) in local model-agnostic explanations. Anchors determine the sufficient part of the input for a specific prediction.       
The presented approach in~\cite{DMM:2018}, named Dirichlet process mixture model with multiple elastic nets (DMM-MEN), also non-linearly approximates the model's decision boundary through a Bayesian model with combination of multiple elastic nets. Hence, DMM-MEN discovers more generalized insights of a model and make a superior explanation for each prediction.      
 
Another recently proposed explanation, named contrastive explanations method (CEM)~\cite{cem}, provides explanation by finding the minimum features in the input which are sufficient to result the same prediction (i.e., Pertinent Positive) accompanied with the minimum features which should be absent in the input to prevent the final prediction change (i.e., Pertinent Negative). For an input image $\textbf{\textit{x}}$, CEM determines Pertinent Positives ($\delta^{pos}$) such that $\operatorname{argmax}_i[Pred(x)]_i=\operatorname{argmax}_i[Pred(\delta)]_i$. Where $\delta \subseteq x$, and $[Pred(x)]_i$ is the $i^{th}$ class prediction for input $\textbf{\textit{x}}$.  
Pertinent Negatives ($\delta^{neg}$) are also determined such that $\operatorname{argmax}_i[Pred(x)]_i=\operatorname{argmax}_i[Pred(x+\delta)]_i$. Here, $\delta\nsubseteq x$. Moreover, three regularizers have been used in minimization process for efficient feature selection, and having a close data manifold to the input.            
\subsection{Decision Analysis}
In this sub-section, we analyze internal component of neural networks for extended transparency of the learned decisions. In particular, one shortcoming of functional methods (Sub-section~\ref{sec:FA}) is that they can not show which neurons play more important role in making a decision. To address this shortcoming, some methods such ~\cite{pred_diff:2017} has generalized their functional approach (\ref{sec:FA}) to the hidden layers. Practically, for a neuron in layer $h_i$, its activation is calculated in two cases: with existing an intended neuron in previous layer $h_{i-1}$ and without it. The difference between them is considered as contribution of the neuron in $h_{i-1}$ on the decision made by the neuron in subsequence layer. 
Moreover, there are some methods which try to explain network's predictions by tracing back and decomposing them through that network. Going backward and constructing a final relevance path can be done using redistribution rules.
One of such techniques is LRP~\cite{LRP:2015} and deep Taylor decomposition in general~\cite{dtd:2017} which is applied to the pre-trained model and redistributes its output iteratively backward through the network using a local propagation rule derived from deep Taylor decomposition (Eq.~\eqref{r2}). Where $R_i^{(l)}$ is the relevance score of neuron $\textbf{\textit{i}}$ in layer $\textbf{\textit{l}}$, and $a_iw_{ij}$ is the input of neuron $\textbf{\textit{j}}$ from neuron $\textbf{\textit{i}}$. Here, the network output is expressed by sum of all neurons' relevance scores in each layer (Eq.~\eqref{r1}). Final pixel relevances depict a heatmap, which visualizes contribution of each pixel.

\begin{equation}\label{r2}
R_i^{(l)}=\sum_{j}\dfrac{a_i w_{ij}}{\sum_{i}a_{i}w_{ij}}R_j^{(l+1)}
\end{equation}   

\begin{equation}\label{r1}
f(I)=...=\sum_{d\in l+1}R_d^{(l+1)}=\sum_{d\in l}R_d^{(l)}=...=\sum_{d\in 1}R_d^{(1)}
\end{equation}     
The naive propagation rule (Eq.~\eqref{r2}) is not numerically stable ($R_i$ can take unbounded large values when $\sum_{i}a_{i}w_{ij}$ is a small value close to zero). Hence, two other LRP rules, named $\epsilon$-rule (Eq.~\eqref{r3}) and $\alpha\beta$-rule (Eq.~\eqref{r4}) have been proposed to avoid unboundedness.

\begin{equation}\label{r3}
R_i^{(l)}=\sum_{j}\dfrac{a_i w_{ij}}{\sum_{i}a_{i}w_{ij}+\epsilon \hspace{.05 in} sign (\sum_{i}a_{i}w_{ij})}R_j^{(l+1)},
\end{equation} 

\begin{equation}\label{r4}
R_i^{(l)}=\sum_{j}(\alpha\dfrac{a_i w_{ij}^+}{\sum_{i}a_{i}w_{ij}^+}+\beta\dfrac{a_i w_{ij}^-}{\sum_{i}a_{i}w_{ij}^-})R_j^{(l+1)},
\end{equation}
where $\epsilon\geq0$ is a stablizer term. $"+"$ and $"-"$ indicates the positive and negative parts respectively. $\alpha$ and $\beta$ are controlling parameters such that $\alpha+\beta=1$.    
Another similar work in this line is called DeepLIFT~\cite{deeplift:2017}. It  propagates important scores according to difference of each neuron's activation with its reference activation. Here, reference activation is the neuron's activation when the network is fed by a reference input (e.g. zero image for MNIST because of all black backgrounds). DeepLIFT helps information could be propagated even in zero gradient condition.         

 

Paper~\cite{koh:2017} explains network's predictions from the training data perspective. 
Determining the most responsible data points (for a specific prediction) by removing each of them and retraining the network is so slow and costly. Therefore, according to Eq.~\eqref{func} each training data $z_i$ is unweighted by a small value $\epsilon$ and its effect on models' parameters is computed using influence functions.  
\newcommand{\defeq}{\overset{\text{def}}{=}}
\begin{equation}\label{func}
I_{up,params}(z)\defeq \dfrac{d\hat{\theta}_{\epsilon,z}}{d\epsilon}\biggm|_{\epsilon = 0}=-H_{\hat{\theta}}^{-1}\nabla_\theta L(z,\hat{\theta})
\end{equation}
\begin{align*}
s.t.\quad \hat{\theta}_{\epsilon,z}=\operatorname*{argmin}_{\theta \in \Theta}\frac{1}{n}\sum_{i=1}^{n}L(z_i,\theta)+\epsilon L(z,\theta),
\end{align*}
\vspace{.05 in}
where $L$ is lost function and $H$ is Hessian matrix. In this view, influence functions can be also used for making adversarial examples to fool the network.
Due to the fact that positive and negative important examples are not provided by influence functions,
a real-time scalable method~\cite{pn:2018} has been proposed to address both positive and negative most responsible training data. Here, model's prediction for a test image is decomposed into the weighted linear combination of influential training points. If large weight is positive, the corresponding point increases the prediction. Otherwise, it belongs to a inhibitory negative point and decreases the prediction. 


%

%
%
%

\section{Explainability by Design} \label{Exp_des}
While existing works mostly focus on explaining the decision-making and contributing factors of established DNNs, building such a posterior knowledge can be both challenging and costly when the architectures and algorithms are not created with explainability in mind. Therefore, it is worth to include explainability into the design of DNNs.

\begin{figure}[t!]
\centering
\includegraphics[width=0.5\textwidth]{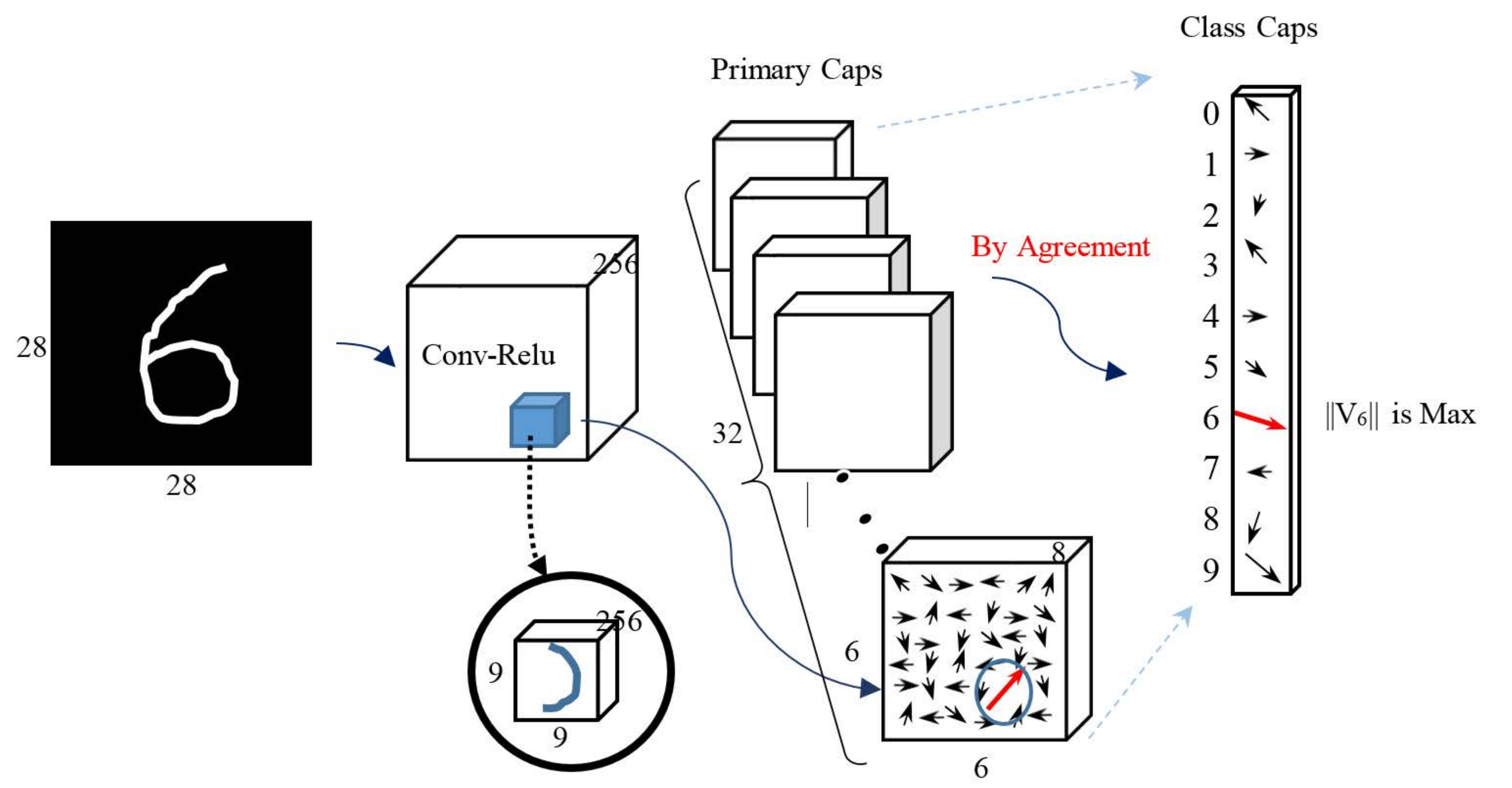}
\caption{\footnotesize Detection architecture of a three layers CapsNet.
In the Primary-Caps layer, each arrow represents activity vector of a capsule. A sample activated capsule has been shown by the red arrow which has higher magnitude for the example kernel introduced to find curve fragment of digit $6$
}
\label{fig:caps}
\end{figure}
The recent adoption of privacy-by-design~\cite{Ann:2009}, i.e., the 
General Data Protection Regulation (GDPR) by the European Union in 2018, has also make it timely and of paramount urgency to dive into explaining decisions made by DNNs.

%
This section investigates some recently proposed DNNs with explainable designs. Capsule Network (CapsNet)~\cite{Caps:2017} is one of those architectures which has been designed with intrinsically explainable structure. CapsNet proposes the idea of equivalence instead of invariance and encapsulating pose information (such as scaling and rotation) and other instantiation parameters using capsules' activity vectors.
Paper~\cite{CapsEx:2018} analyzes CapsNet's nested architecture (Fig.~\ref{fig:caps}) and verifies its explainable properties. Each capsule consists of a likelihood value and an instantiation parameter vector. For an object capsule, when all its component capsules' vectors in previous layer are consistent together (e.g., all of them are in the same scale and rotation), it will have higher likelihood which explains high probability of the feature that the capsule detects. Moreover, during learning procedure (routing-by-agreement) unrelated capsules become smaller through non-squashing function and resulting coupling coefficients. Hence, CapsNet intrinsically construct a relevance path without applying a backward process for explanation.

\renewcommand{\arraystretch}{2.0}
\begin{table*}[b]
	\centering
	\caption{ Summary of DNNs understanding, visualization an explanation methods}
	\setlength{\tabcolsep}{19pt}

	\begin{tabular*}{\textwidth}{ @{\extracolsep{\fill}}cccc}

		\hline
		\toprule
		Reference & Year & Category & Method \\
		\hline
		\midrule
		\rowcolor{pink!20}
		\cite{ImgN:2014} & 2014 & CNNs/First Layer & Visualizing weights of conv filters in the first layer \\	
		\cite{I2}& 2014 & \hspace{0.7cm} CNNs/Intermediate Layer\hspace{0.7cm}  & \hspace{1cm}Finding input patches which maximize a neuron activation\hspace{0.5cm}\\
		\rowcolor{pink!20}
		\cite{I8}& 2014 & CNNs/Intermediate Layer  & Visualizing through activation maximization\\
		\cite{I1}& 2015 & CNNs/Intermediate Layer & Synthesizing the input which maximizes a neuron activation   \\
		\rowcolor{pink!20}
		\cite{I6-1}& 2016 & CNNs/Intermediate Layer & Back-propagating output weights to the last Conv layer  \\
		\cite{I6}& 2016 & CNNs/Intermediate Layer & Back-propagating output weights to the last Conv layer  \\
		\rowcolor{pink!20}
		\cite{I10}& 2016 & CNNs/Intermediate Layer & Multifaceted feature visualization  \\
		\cite{I11}& 2016 & CNNs/Intermediate Layer & Synthesizing the input which maximizes a neuron activation 

	\end{tabular*}
	 \label{tab:summery}
\end{table*}
Another architecture is an interpretable feed-forward design for CNNs (CNN/FF) rather than the original CNNs with back-propagation~\cite{ff:2018}. Each convolutional layer in CNN/FF is considered as a vector space transformation and filter weights are computed using the introduced subspace approximation with adjusted bias (Saab) transform which is a PCA-based transformation. 
Moreover, in each fully-connected layer, CNN/FF applies k-means clustering on the input and combines the class and cluster labels as $k$ pseudo-labels for that input. Then, a linear least square regression (LSR) is computed by these pseudo-labels and the output is obtained accordingly.   
Since CNN/FF is based on linear algebra and data statistics, its structure is mathematically explainable. 

Paper~\cite{excnn:2017} also proposes an interpretable CNN with semantical object-part filters in high convolutional layers. Since each of these filters is activated by a specific part of an object rather than activating by mixture of patterns in common CNNs, there is a disentangled representation in these convolutional layers. Here, a mutual-information-based loss is defined for each filter which enforces it to capture a more distinct object part of a category in an unsupervised manner. The authors also improved their approach by measuring each object-part filter's contribution in final decisions and learning the corresponding decision tree which explains network's predictions quantitatively~\cite{tree:2018}.

\begin{figure} 
	\begin{tikzpicture}
	\begin{axis}[
	width=8.5cm,
	height=10cm,
	ymajorgrids=true,
	grid style=dashed,
	legend style={at={(0.5,-0.15)},
		anchor=north,legend columns=-1},
	ylabel={Number of publications},
	symbolic x coords={2010, 2011, 2012, 2013, 2014, 2015, 2016, 2017, 2018},
	xtick=data,
	x tick label style={rotate=45,anchor=east},
	nodes near coords align={vertical},
	]
	\addplot[ybar, fill=blue!50] 
	coordinates {(2010,46) (2011,55) (2012,56) (2013,56) (2014,75) (2015,74) (2016,110) (2017,264) (2018,610)};
	\addplot[draw=red,ultra thick,smooth,] 
	coordinates {(2010,46) (2011,55) (2012,56) (2013,56) (2014,75) (2015,74) (2016,110) (2017,264) (2018,610)};
	\end{axis}
	\end{tikzpicture}
	\caption{\footnotesize Increasing interests on explainability, interpretability and visualization of DNNs based on Google scholar data.}
	\label{fig:cite}	
\end{figure}
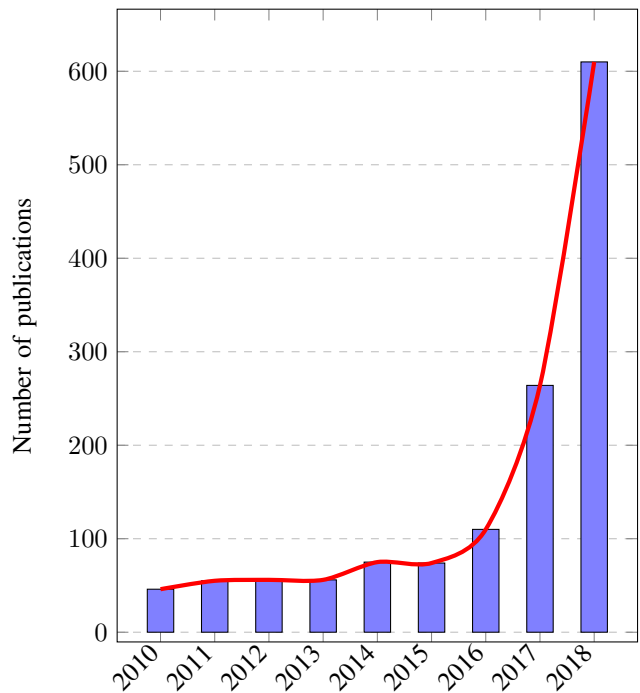

\section{Conclusion and Future Works} \label{sec:Conc}
Increasing interests on Explainable Artificial Intelligence (XAI), as displayed in Fig.~\ref{fig:cite}, has motivated scientists to do more research in this area.
In this article, we presented the necessity of explanation, visualization, and understanding of Deep Neural Networks (DNNs), specially in complex machine learning and critical computer vision tasks. We advanced our knowledge in this domain by reviewing the state-of-the-art interpreting and explaining solutions in three main categories (structural analysis, behavioral analysis and explainability by design).
Table~\ref{tab:summery} presents a summarization of all the reviewed methods. However, There are still three key items that require further investigation as briefly outlined below:

1. Since there are some visual artifacts in image data, which can be wrongly considered as a kind of explanation or mislead interpreting process, it is critical to evaluate and validate the obtained explanations using a sanity check testing. Additionally, a criteria is required for comparing different explanations and choosing the best one to trust a model.

2. Interpreting and explaining Deep Generative Models (DGMs), as the most promising unsupervised and semi-supervised learning approaches, has not completely investigated before. DGMs aim to learn a model that represents true data distribution of training samples and then generate real samples from that distribution. The most popular DGMs include generative adversarial networks (GAN) and variational autoencoders (VAE) which it is important to analyze their structure through understanding the latent space and learning disentangled representations. 
 
3. Due to the advantages of embedding explainability in architecture and few existing such methods, there is still a room for designing more inherently explainable strategies for DNN structures.

4. Adversarial examples are the inputs which cause a model makes an incorrect prediction. Considering that explanations discover the reasons behind a prediction, they can be used to improve DNNs resistance against various adversarial attacks. 
\\

\renewcommand{\arraystretch}{2.0}
\begin{table*}[t]
	\centering
	\setlength{\tabcolsep}{22pt}
	\begin{tabular*}{\textwidth}{ @{\extracolsep{\fill}}cccc}
		
		\rowcolor{pink!20}
		\cite{I4}& 2015 & CNNs/Intermediate Layer& Feature inversion\\
		
		\cite{I5}& 2016 & CNNs/Intermediate Layer& 
		Feature inversion\\
		\rowcolor{pink!20}
		\cite{style2:2016}& 2016 & CNNs/Intermediate Layer& 
		Image super-resolution and Style transfer\\
		
		\cite{texture:2015}& 2015 & CNNs/Intermediate Layer & 
		Texture synthesis\\
		\rowcolor{pink!20}
		\cite{style:2016}& 2016 & CNNs/Intermediate Layer  & 
		Style transfer\\
		\cite{style3:2016}& 2016 & CNNs/Intermediate Layer & 
		Style transfer\\
		\rowcolor{pink!20}
		\cite{style5:2016}& 2016 & CNNs/Intermediate Layer  & 
		Style transfer\\
		\cite{style6:2017}& 2017 & CNNs/Intermediate Layer & 
		Style transfer\\
		\rowcolor{pink!20}
		\cite{I9}& 2014 & CNNs/First and Intermediate Layers & 
		Generality and specificity concepts of conv features\\
		\cite{ImgN:2014}& 2014 & CNNs/Last Layer & Visualizing last layer feature vectors by nearest neighbor 
		\\
		\rowcolor{pink!20}
		\cite{L2}& 2008 & CNNs/Last Layer  & 
		Visualizing last layer feature vectors by t-SNE\\
		\cite{RNN1}& 2015 & RNNs & 
		Static visualization for character-level language modeling\\
		\rowcolor{pink!20}
		\cite{RNN_lstmvis}& 2016 & RNNs 
		& An interactive tool to identify hidden units dynamics\\
		\cite{RNN_NLP}& 2015 & RNNs & 
		Gradient-based saliency technique for visualizing local semantics\\
		\rowcolor{pink!20}
		\cite{RNN2}& 2017 & RNNs & 
		Modeling the relation between hidden units and inputs
\\
		\cite{RNN_attention}& 2014 & RNNs & 
		Attention mechanism for finding more relevant phrases\\
		\rowcolor{pink!20}
		\cite{RNN_attention2}& 2017 & RNNs & 
		Quantifying contribution of each word in final representation\\
		\cite{RNN_erasure}& 2016 & RNNs & 
		Monitoring the effects of removing items in final decision\\
		\rowcolor{pink!20}
		\cite{I2}& 2014 & Functional Analysis & 
		Sensitivity analysis\\
		\cite{pred_diff:2017}& 2017 & Functional and Decision Analysis& 
		Sensitivity analysis\\
		\rowcolor{pink!20}
		\cite{SA:2010}& 2010 & Functional Analysis& 
		Determining pivotal pixels in classifier decisions\\
		\cite{I3}& 2013 & Functional Analysis& 
		Determining pivotal pixels in classifier decisions\\
		\rowcolor{pink!20}
		\cite{smoothgrad}& 2017 & Functional Analysis & 
		Smooth sensitivity map\\
		\cite{IG}& 2016 & Functional Analysis& 
		Integrated gradients of scaled copies of original input\\
		\rowcolor{pink!20}
		\cite{Marco:2016}& 2016 & Functional Analysis & 
		LIME\\
		\cite{anchor:2018}& 2018 & Functional Analysis & 
		Anchor\\
		\rowcolor{pink!20}
		\cite{DMM:2018}& 2018 & Functional Analysis& 
		DMM-MEN\\
		\cite{cem}& 2018 & Functional Analysis& 
		CEM\\
		\rowcolor{pink!20}
		\cite{LRP:2015}& 2015 & Decision Analysis& 
		LRP\\
		\cite{deeplift:2017}& 2017 & Decision Analysis& 
		DeepLIFT\\
		\rowcolor{pink!20}
		\cite{koh:2017}& 2017 &  Decision Analysis& 
		Explaining decisions from training data perspective\\
		\cite{pn:2018}& 2018 & Decision Analysis& 
		 Explaining decisions from training data perspective\\
		 \rowcolor{pink!20}
		\cite{CapsEx:2018}& 2018 & 
		 Explainability by design & 
		CapsNet\\
		\cite{ff:2018}& 2018 &  Explainability by design & 
		CNN/FF
		\\
		\rowcolor{pink!20}
		\cite{excnn:2017}& 2017 & 
		Explainability by design & 
		Semantical object-part filters\\
		\cite{tree:2018}& 2018 & Explainability by design & 
		Semantical object-part filters decision tree
		\\
		\hline
	\end{tabular*}
\end{table*}

\FloatBarrier
\bibliographystyle{IEEEbib}

\small
\vspace{.1in}
\noindent

\end{document}